%% file: main.tex
\documentclass[10pt]{article} 
\usepackage[accepted]{rlc}

\usepackage{amssymb}            
\usepackage{mathtools}          
\usepackage{mathrsfs}           
\usepackage{graphicx}           
\usepackage{subcaption}         
\usepackage[space]{grffile}     
\usepackage{url}                

\title{Demystifying the Recency Heuristic in\\Temporal-Difference Learning}

\author{Brett Daley\\
    brett.daley@ualberta.ca\\
    Dept.\ of Computing Science\\
    University of Alberta\\
    \And
    Marlos C. Machado\\
    machado@ualberta.ca\\
    Dept.\ of Computing Science\\
    University of Alberta\\
    Canada CIFAR AI Chair
    \And
    Martha White\\
    whitem@ualberta.ca\\
    Dept.\ of Computing Science\\
    University of Alberta\\
    Canada CIFAR AI Chair
}

\input{math_commands.tex}

\usepackage{paper_style}
\graphicspath{{images/}}

\begin{document}

\maketitle

\input{paper_body.tex}

\hfill

\subsubsection*{Acknowledgments}
We thank Rich Sutton for helpful discussions and insights.
This research is supported in part by the Alberta Machine Intelligence Institute (Amii), the Natural Sciences and Engineering Research Council of Canada (NSERC), and the Canada CIFAR AI Chair Program.

\clearpage


\bibliography{main}
\bibliographystyle{rlc}

\appendix

\input{paper_appendix.tex}

\end{document}

%% file: math_commands.tex
\usepackage{amsmath,amsfonts,bm}

\def\eqref#1{equation~\ref{#1}}

\def\1{\bm{1}}

\def\vc{{\bm{c}}}

\def\vq{{\bm{q}}}
\def\vr{{\bm{r}}}

\def\vv{{\bm{v}}}
\def\vw{{\bm{w}}}
\def\vx{{\bm{x}}}

\def\mH{{\bm{H}}}

\def\mP{{\bm{P}}}

\def\mT{{\bm{T}}}

\def\mX{{\bm{X}}}

\def\mPi{{\bm{\Pi}}}

\DeclareMathAlphabet{\mathsfit}{\encodingdefault}{\sfdefault}{m}{sl}
\SetMathAlphabet{\mathsfit}{bold}{\encodingdefault}{\sfdefault}{bx}{n}

\newcommand{\Var}{\mathrm{Var}}

\newcommand{\Cov}{\mathrm{Cov}}



%% file: paper_body.tex
\begin{abstract}
    \noindent
    The recency heuristic in reinforcement learning is the assumption that stimuli that occurred closer in time to an acquired reward should be more heavily reinforced.
    The recency heuristic is one of the key assumptions made by TD($\lambda$), which reinforces recent experiences according to an exponentially decaying weighting.
    In fact, all other widely used return estimators for TD learning, such as $n$-step returns, satisfy a weaker (i.e., non-monotonic) recency heuristic.
    Why is the recency heuristic effective for temporal credit assignment?
    What happens when credit is assigned in a way that violates this heuristic?
    In this paper, we analyze the specific mathematical implications of adopting the recency heuristic in TD learning.
    We prove that any return estimator satisfying this heuristic:
    1)~is guaranteed to converge to the correct value function,
    2)~has a relatively fast contraction rate, and
    3)~has a long window of effective credit assignment, yet bounded worst-case variance.
    We also give a counterexample where on-policy, tabular TD methods violating the recency heuristic diverge.
    Our results offer some of the first theoretical evidence that credit assignment based on the recency heuristic facilitates learning.
    \looseness=-1
\end{abstract}

\section{Introduction}

The temporal credit-assignment problem in reinforcement learning (RL) is the challenge of determining which past actions taken by a decision-making agent contributed to a certain outcome \citep{minsky1961steps}.
Addressing the temporal credit-assignment problem effectively is paramount to efficient RL.
Unfortunately, an optimal solution is likely infeasible for an agent acting in an arbitrary, unknown environment;
perfect credit assignment would require precise knowledge of the environment's dynamics.
Even then, the complexity of the problem grows enormously as the agent takes more actions over its lifetime.
Instead, heuristics---simplifying rules or assumptions for credit assignment---can be adopted to make the problem more approachable.
In the absence of any prior knowledge of the environment, a common and reasonable choice is the \emph{recency heuristic}:
``One assigns credit for current reinforcement to past actions according to how recently they were made'' \citep[][p.~94]{sutton1984temporal}.
The recency heuristic reflects the fact that there is likely to be a cause-and-effect relationship between actions and rewards that are close together in time.

In computational RL, the reinforcement signal is taken to be the temporal-difference (TD) error:
the difference between the observed and expected reward earned by an action.
TD($\lambda$) \citep{sutton1988learning} is the prime example of the recency heuristic;
each TD error is applied to past actions in proportion to an exponentially decaying eligibility, achieving credit assignment that gracefully fades as the time between the action and TD error increases.
This strategy, although simple, is highly effective and has been used by many recent algorithms \citep[e.g.,][]{schulman2015high,harb2016investigating,harutyunyan2016q,munos2016safe,vanseijen2016effective,mahmood2017multi,mousavi2017applying,daley2019reconciling,kozuno2021revisiting,gupta2023past,tang2024off}.
\looseness=-1

\begin{figure}
    \centering
    \includegraphics[width=\textwidth]{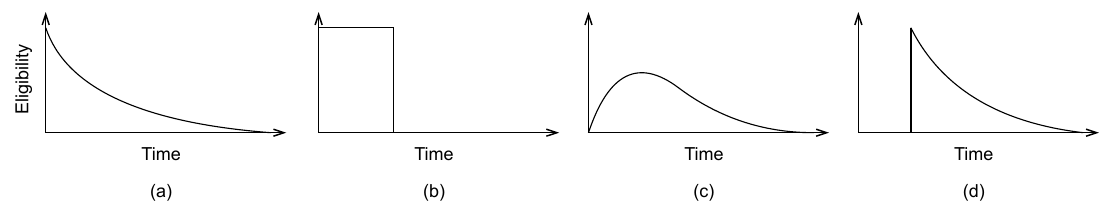}
    \caption{Illustrations of the eligibility curves for (a) $\lambda$-return, (b) $n$-step return, (c) inverted U-shape assignment inspired by \citet{klopf1972brain}, and (d) time-delayed $\lambda$-return.
    The horizontal axis represents the elapsed time since the stimulus.
    Neither (c) nor (d) satisfy the recency heuristic.}
    \label{fig:curves}
\end{figure}

However, the recency heuristic is, by definition, a simplifying assumption;
one can imagine complex environments where non-recent credit assignment would theoretically be more beneficial.
For example, if it were known that there is always some fixed delay between actions and their corresponding effects---especially when under partial observability \citep{kaelbling1998planning}---then this information could theoretically be exploited for faster learning.
\citet{klopf1972brain}, for instance, describes credit-assignment functions based on an inverted-U shape (see \Cref{fig:curves}c) that could achieve this exact effect.
The shape of the credit-assignment curve encodes a prior belief over the likelihood of when a reward will arrive following an action, with the smooth distribution reflecting some uncertainty in the exact time of arrival.
\citet{klopf1972brain} hypothesized that reactions in a firing neuron would leave it eligible to learn for a short duration.
This later inspired the simplified spike-and-decay model of eligibility traces \citep{barto1983neuronlike,sutton1984temporal} used by TD($\lambda$), which obeys the recency heuristic and has become a standard approach for credit assignment in computational RL.

Although there is potential for more efficient learning with non-recent credit assignment, it has not been tried in computational RL.
Even alternatives to TD($\lambda$) that are not generally connoted with the recency heuristic, such as $n$-step TD methods \citep{cichosz1995truncating}, implement a crude form of recency heuristic:
TD errors within some fixed time interval following an action are reinforced, while those outside are not.
In fact, all other return estimators used for TD learning (which are constructed from $n$-step returns) satisfy some form of recency heuristic (see \Cref{sec:convex_wrh}).
We are not aware of any results that analyze what happens when TD updates do not follow the recency heuristic.

The goals of this paper are to understand the implications of forgoing the recency heuristic in TD learning, and to provide new insights into why assigning credit based on the recency heuristic has been so effective for RL.
We test a model of non-recent credit assignment based on a short, time-delayed pulse inspired by \citeauthor{klopf1972brain}'s \citeyearpar{klopf1972brain} inverted-U function.
Although this is one of the simplest and most benign forms of non-recency in TD learning, we show that it diverges under the favorable conditions of tabular, on-policy learning.
We prove that the root cause of divergence is negative weights on some of the $n$-step returns in the return estimate, which appear whenever the recency heuristic is violated, and counteract learning by increasing the contraction modulus.
In the off-policy setting, our analysis resolves the open problem by \citet{daley2023trajectory} on the convergence of trajectory-aware eligibility traces.
Finally, we show that satisfying the recency heuristic increases the effective credit-assignment window of a return estimate without increasing its bias and variance in the worst case, which partly explains the empirical success of methods like TD($\lambda$).
Overall, our results demonstrate that the recency heuristic is not an overly simplistic assumption but is actually a crucial component in the mathematical basis of TD learning.

\section{Background}
\label{sec:bg}

We adopt the standard RL perspective of a decision-making agent learning in an unknown environment through trial and error \citep[][Sec.~3.1]{sutton2018reinforcement}.
The agent-environment interface is modeled by a Markov decision process (MDP) formally described by the tuple $(\mathcal{S},\mathcal{A}, p, r)$.
The finite sets $\mathcal{S}$ and $\mathcal{A}$ contain the possible environment states and agent actions, respectively.
At each time step $t \geq 0$, the agent observes the current state of the environment, $S_t$, and takes an action, $A_t \in \mathcal{A}$, with probability $\pi(A_t|S_t)$, where $\pi$ is the agent's policy.
Consequently, the environment state transitions to $S_{t+1} \in \mathcal{S}$ with probability $p(S_{t+1}|S_t,A_t)$, and the agent receives a reward, $\smash{R_t \defeq r(S_t,A_t)}$.
\looseness=-1

In prediction problems, the agent's objective is to learn the value function $v_\pi(s) \defeq \mathbb{E}_\pi[G_t \mid S_t=s]$, where
$\smash{G_t \defeq \sum_{i=0}^\infty \gamma^i R_{t+i}}$
is the observed discounted return.
The constant $\gamma \in [0,1]$ is called the discount factor and determines the agent's relative preference for delay rewards.
In the rest of this section, we discuss various types of temporal-difference (TD) learning \citep{sutton1988learning}, a common approach for prediction in reinforcement learning.

\paragraph{TD($\lambda$) and the Recency Heuristic}

TD methods estimate $v_\pi$ by iteratively reducing an error between predicted and observed returns, \emph{bootstrapping} from the previous (biased) estimates in order to reduce variance.
Let $v \colon \mathcal{S} \to \mathcal{A}$ be the agent's estimate of the value function, and define $\smash{V_t \defeq v(S_t)}$ for brevity.
The TD error, defined as $\smash{\delta_t \defeq R_t + \gamma V_{t+1} - V_t}$, is the fundamental unit of reinforcement in TD methods.
For instance, the simplest TD method, known as TD(0) or $1$-step TD \citep{sutton1988learning}, performs the update
$v(S_t) \gets V_t + \alpha_t \delta_t$, where $\alpha_t \in (0,1]$ is the step size.
TD(0) is a special case of TD($\lambda$) \citep{sutton1988learning}, one of the earliest and most widely used TD methods.
TD($\lambda$) is able to assign credit simultaneously to multiple states through the use of eligibility traces \citep{klopf1972brain,barto1983neuronlike,sutton1984temporal},
a function $z \colon \mathcal{S} \to \mathbb{R}$ that tracks recent state visitations.
On each time step, TD($\lambda$) performs the following updates:
\begin{align}
    \label{eq:etraces}
    z(s) \gets \gamma \lambda \mathop{z(s)}\,,\ \forall~s \in \mathcal{S}\,, &&
    z(S_t) \gets z(S_t) + 1 \,, &&
    v(s) &\gets v(s) + \alpha_t \delta_t \mathop{z(s)}
    \,,\
    \forall~s \in \mathcal{S}
    \,,
\end{align}
where $\lambda \in [0,1]$ is the recency hyperparameter.
Every eligibility trace is unconditionally decayed by a factor of $\gamma \lambda$, but only the trace for the current state is incremented.
Then, every state is updated in proportion to its eligibility trace, using the current TD error.
Eligibility traces are an efficient mechanism for assigning credit to recently visited states.

The above updates are known as the \emph{backward view} of TD($\lambda$).
An alternative perspective is the \emph{forward view}.
Suppose we hold the value function and step size fixed, and track the cumulative update for a single state visitation.
We would find that the state is updated according to
\begin{align}
    \label{eq:tdlambda_forward}
    &v(S_t) \gets V_t + \alpha_t \big( G^\lambda_t - V_t)
    \,, \\
    \label{eq:lambda-return_errors}
    &\text{where } G^\lambda_t \defeq V_t + \sum_{i=0}^\infty (\gamma \lambda)^i \delta_{t+i}
    \,.
\end{align}
The forward and backward views are equivalent under the conditions described above \citep{sutton1988learning,watkins1989learning}.
The quantity defined in \Cref{eq:lambda-return_errors} is known as the $\lambda$-return and represents the theoretical target of the TD($\lambda$) update.
Although the forward view is acausal and not directly implementable as an online algorithm, it reveals the temporal relationship between a state and the degree to which future TD errors are reinforced.
The exponential decay of \Cref{eq:lambda-return_errors} represents a form of \emph{recency heuristic}, the assumption that the causality between events weakens as the time between them increases.
Mathematically, the hyperparameter $\lambda$ controls the bias-variance trade-off by interpolating between high-bias $1$-step TD (${\lambda=0}$) and high-variance Monte Carlo (${\lambda=1}$) methods \citep{kearns2000bias}.
As we show next, non-exponential implementations of the recency heuristic are also possible;
however, they do not enjoy the same efficient implementation with eligibility traces.

\paragraph{$n$-step Returns and Compound Returns}

More generally, TD methods can be expressed as a forward-view update in terms of an arbitrary return estimate, $\hat{G}_t$:
\begin{equation}
    \label{eq:backup}
    v(S_t) \gets V_t + \alpha_t \big( \hat{G}_t - V_t \big)
    \,.
\end{equation}
This operation is known as a value backup, and we refer to the estimate $\hat{G}_t$ as its target.
We already established in \Cref{eq:tdlambda_forward} that the $\lambda$-return, $G^\lambda_t$, is one possible target.
Another common target is the $n$-step return \citep{watkins1989learning,cichosz1995truncating}, defined as
$\smash{\nstep{n}_t \defeq \sum_{i=0}^{n-1} \gamma^i R_{t+i} + \gamma^n V_{t+n}}$,
where $n \geq 1$ determines the length of the return.
Just like the $\lambda$-return, the $n$-step return interpolates between high-bias TD ($n=1$) and high-variance Monte Carlo ($n=\infty$) methods.
Although not commonly used, the $n$-step return admits a forward-view cumulative error similar to \Cref{eq:lambda-return_errors}:
\begin{equation}
    \label{eq:nstep-return_errors}
    \nstep{n}_t = V_t + \sum_{i=0}^{n-1} \gamma^i \delta_{t+i}
    \,,
\end{equation}
This reveals that the $n$-step return also satisfies the recency heuristic, albeit a weaker notion than that of the $\lambda$-return (see \Cref{sec:recency_heuristic}).
Nevertheless, it still fulfills the basic assumption that TD errors nearer in time to a given state should be reinforced, whereas those farther away should not.
The $n$-step return is also useful as a fundamental building block for constructing other estimates.
For instance, the $\lambda$-return from \Cref{eq:lambda-return_errors} is equivalent to a weighted average of $n$-step returns:
\begin{equation}
    \label{eq:lambda-return_nsteps}
    G^\lambda_t = (1-\lambda) \sum_{n=1}^\infty \lambda^{n-1} \nstep{n}_t
    \,.
\end{equation}
More generally, we can consider arbitrary convex combinations of $n$-step returns, strictly generalizing both $\lambda$-returns and $n$-step returns.
Let $(c_n)_{n=1}^\infty$ be a sequence of nonnegative weights such that $\sum_{n=1}^\infty c_n = 1$.
We refer to the following estimate as a \emph{convex} return:
\begin{equation}
    \label{eq:convex_return}
    G^\vc_t \defeq \sum_{n=1}^\infty c_n \nstep{n}_t
    \,.
\end{equation}
When at least two weights are nonzero, a convex return becomes a weighted average of $n$-step returns known as a compound return \citep{watkins1989learning,sutton2018reinforcement,daley2024averaging}.
Examples of compound returns include $\lambda$-returns, $\gamma$-returns \citep{konidaris2011td_gamma}, and $\Omega$-returns \citep{thomas2015policy}.
In \Cref{sec:convex_wrh}, we show that the definition of a convex return is inherently related to the recency heuristic.
Prior to our work, convex returns were the most general form of return estimator for TD learning, but we generalize them further in \Cref{sec:convex_wrh}.

\paragraph{Value-Function Operators and Convergence Conditions}

We have discussed a variety of TD methods based on forward-view return estimates, but we have not yet established what makes an estimate valid for learning.
Convergence to $v_\pi$ is perhaps most easily seen from the perspective of value-function operators.
An operator $\mH \colon \mathbb{R}^{\abs{\mathcal{S}}} \to \mathbb{R}^{\abs{\mathcal{S}}}$ transforms a value function.
The most fundamental value-function operator is the Bellman operator \citep{bellman1957dynamic}, defined as
\begin{equation*}
    \mT_\pi \vv \defeq \vr + \gamma \mP_\pi \vv
    \,,\enskip
    \text{where } (\mP_\pi \vv)(s) \defeq \sum_{a \in \mathcal{A}} \pi(a|s) \sum_{s' \in \mathcal{S}} \mathop{p(s'|s,a)} \vv(s')
    \,.
\end{equation*}
Note that $\vr$ and $\vv$ here are treated as vectors in $\mathbb{R}^{\abs{\mathcal{S}}}$, and $\mP_\pi$ is treated as a $\abs{\mathcal{S}} \times \abs{\mathcal{S}}$ stochastic matrix.
Let
$\smash{\mT_\pi^n \vv \defeq \mT_\pi \mT_\pi^{n-1} \vv}$
and
$\smash{\mT_\pi^0 \vv \defeq \vv}$.
The $n$-iterated Bellman operator, $\mT_\pi^n$, corresponds to the $n$-step return.
Hence, convex returns are associated with the operator $\vv \mapsto \sum_{n=1}^\infty c_n \mT_\pi^n \vv$.
More generally, every value backup like \Cref{eq:backup} is equivalent to the noisy application of some operator, $\mH$, to an element of the value function.
That is, a return estimate can be represented as
${\hat{G}_t = (\mH \vv)(S_t) + \omega_t}$,
where $\omega_t$ is zero-mean noise.
TD updates can thus be expressed in the form
\begin{equation}
    \label{eq:op_update}
    \vv(s) \gets \begin{cases}
        (1-\alpha_t) \mathop{\vv(s)} + \alpha_t \big( (\mH \vv)(s) + \omega_t \big) \,, & \text{if } s = S_t \,, \\
        \vv(s) \,, & \text{otherwise.} \\
    \end{cases}
\end{equation}
To produce a TD method of the form of \Cref{eq:backup} that converges to $\vv_\pi$ under general conditions \citep[e.g.,][Proposition~4.4]{bertsekas1996neuro}, it is required that $\mH$ is a maximum-norm contraction mapping with $\vv_\pi$ as its unique fixed point, and that the step sizes are annealed such that
${\sum_{t=0}^\infty \alpha_t = \infty}$ and ${\sum_{t=0}^\infty \alpha_t^2 < \infty}$ \citep{robbins1951stochastic}.
An operator $\mH$ is a contraction mapping if and only if
$\norm{\mH \vv - \mH \vv'}_\infty \leq \beta \norm{\vv - \vv'}_\infty$,
where $\beta \in [0,1)$ is the contraction modulus.
All of the operators discussed so far satisfy these properties because they are convex combinations of $n$-step Bellman operators, each of which are contraction mappings around $\vv_\pi$ with a modulus of $\gamma^n$.
\looseness=-1

\section{Formalizing the Recency Heuristic}
\label{sec:recency_heuristic}

In this section, we precisely define the notion of the recency heuristic.
We consider a general estimator for TD learning of the form
\begin{equation}
    \label{eq:td_error_forward_view}
    \hat{G}_t
    = V_t + \sum_{i=0}^\infty h_i \gamma^i \delta_{t+i}
    \,,
\end{equation}
where $(h_i)_{i=0}^\infty$ is a sequence of real numbers.
Although this may appear to be restrictive, we show in \Cref{sec:convex_wrh} that it can represent every valid return estimate (i.e., converges to $\vv_\pi$) that comprises a linear combination of future rewards and state values, and thus is implementable as a TD method.

We can think of \Cref{eq:td_error_forward_view} as an abstract form of TD($\lambda$):
one with an arbitrary stimulus-response model rather than the familiar exponential decay.
At time~$t$, the agent experiences an external stimulus modulated by the current environment state, $S_t$.
Positive or negative reinforcement subsequently arrives in the form of the TD errors, $(\delta_t, \delta_{t+1}, \delta_{t+2}, \dots)$.
Each weight, $h_i$, determines the agent's receptiveness, or eligibility, to learn from the TD error that occurs exactly $i$ steps after the initial stimulus.
In this view, a return estimate, $\hat{G}_t$, is uniquely determined by the impulse response of a linear time-invariant system encoded by $(h_i)_{i=0}^\infty$.
One possible interpretation of the recency heuristic, then, is a constraint on the impulse response such that it never increases after the initial stimulus.
This gives us the following definition.

\begin{definition}[Weak Recency Heuristic]
    \label{def:wrh}
    A return estimate satisfies the weak recency heuristic if and only if it has the form of \Cref{eq:td_error_forward_view}, and
    $h_i \geq h_{i+1} \geq 0$ holds for all $i \geq 0$.
\end{definition}

We show in \Cref{sec:convex_wrh} that this definition is highly related to the question of whether (and how fast) TD learning using this estimator converges in expectation, but it is slightly weaker than what is typically thought of as the recency heuristic.
For instance, \citet[][p.~94]{sutton1984temporal} is explicit that ``Credit assigned should be a monotonically decreasing function of the time between action and reinforcement, approaching zero as this time approaches infinity.''
The credit-assignment function in \Cref{def:wrh} is merely nonincreasing, and so we refer to it as the \emph{weak} recency heuristic.
Alternatively, we refer to the monotonically decreasing case as the \emph{strong} recency heuristic, defined below.
\looseness=-1

\begin{definition}[Strong Recency Heuristic]
    \label{def:srh}
    A return estimate satisfies the strong recency heuristic if and only if it has the form of \Cref{eq:td_error_forward_view}, and $h_i > h_{i+1} > 0$ holds for all $i \geq 0$.
\end{definition}

Notice that \Cref{def:srh} implies \Cref{def:wrh}.
We make the distinction between these more concrete with a few examples.
The $\lambda$-return, used by TD($\lambda$), is the canonical example of the strong recency heuristic;
its eligibility weights in \Cref{eq:lambda-return_errors} are strictly decreasing for any $\lambda \in (0,1)$.
In contrast, the $n$-step return remains equally receptive to the first $n$ TD errors, and then abruptly stops responding to the ones afterwards.
However, these two updates are alike in that the weights never increase at any point:
they both satisfy the weak recency heuristic.
We could also imagine arbitrary weights in \Cref{eq:td_error_forward_view} that do not satisfy either definition of recency heuristic.
For example, the inverted-U shape described by \citet{klopf1972brain} takes time to reach its peak value before falling back to zero, and thus violates \Cref{def:wrh,def:srh}.
Similarly, we can take the standard spike-and-decay model of a $\lambda$-return and introduce a delay between the initial stimulus and the response.
Both of these could exploit some known structure regarding the agent's environment, and may be more biologically plausible, but their mathematical implications are not yet known.
These four examples are graphed in \Cref{fig:curves}.
Notably, there are many more possibilities in \Cref{eq:td_error_forward_view}, most of which have not yet been explored.

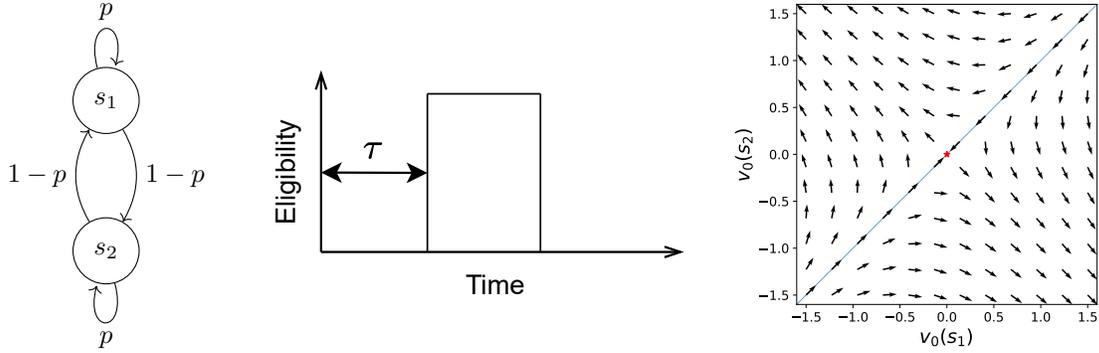
\begin{figure}
    \centering
    \hfill
    \begin{minipage}[b]{0.2\textwidth}
        \vbox to 4cm{
            \begin{tikzpicture}
                \node[state] (s1) {$s_1$};
                \node[state, below of=s1] (s2) {$s_2$};
                \draw (s1) edge[loop above] node{$p$} (s1)
                      (s1) edge[bend left, right] node{$1-p$} (s2)
                      (s2) edge[loop below] node{$p$} (s2)
                      (s2) edge[bend left, left] node{$1-p$} (s1);
            \end{tikzpicture}
        }
    \end{minipage}
    \hfill
    \begin{minipage}[t]{0.4\textwidth}
        \vbox to 3cm{
            \vfill
            \hfill
            \includegraphics[width=0.9\textwidth]{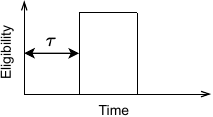}
            \vfill
        }
    \end{minipage}
    \hfill
    \begin{minipage}[b]{0.4\textwidth}
        \vbox to 4cm{
            \centering
            \vfill
            \includegraphics[width=0.8\textwidth]{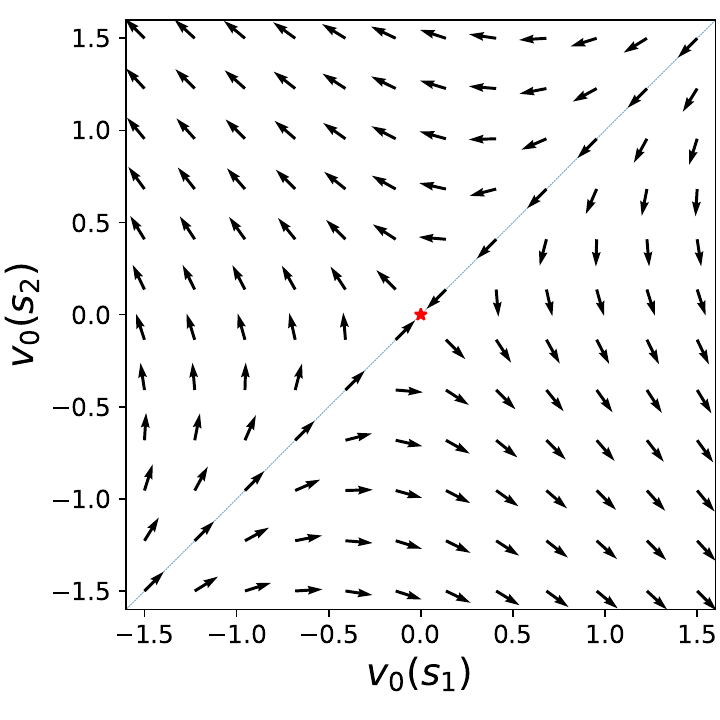}
            \vfill
        }
    \end{minipage}
    \hfill
    \vspace{0.15in}  
    \caption{
        (Left) MRP for \Cref{counterexample:pulse}; rewards are zero.
        (Center) Credit-assignment function for delayed TD(0).
        (Right) Expected update directions of \Cref{eq:delayed_pulse} for $\tau=1$, $\gamma=0.9$, $p=0.4$.
    }
    \label{fig:counterexample}
\end{figure}

\section{What Happens When the Recency Heuristic Is Violated?}
\label{sec:counterexample}

We conduct an experiment to demonstrate that on-policy TD learning with a tabular value function can diverge when the recency heuristic is violated.
This is surprising, since one view of the TD-error weights, $(h_i)_{i=0}^\infty$, is that they encode a belief over the time when rewards will arrive following a stimulus.
Ideally, these weights could represent any shape for the credit-assignment function and the agent would still learn the correct value function, yet this does not appear to be the case.

We test perhaps the simplest possible example of non-recent credit assignment:
an update based on a single, future TD error.
More specifically, we generalize TD(0) by introducing a delay of $\tau \geq 0$:
\begin{equation}
    \label{eq:delayed_pulse}
    v(S_t) \gets V_t + \alpha_t \gamma^\tau \delta_{t+\tau}
    \,.
\end{equation}
The impulse response for this method is generally given by $h_i = 1$ if $i = \tau$, and $h_i = 0$ otherwise.
That is, the eligibility curve is a square pulse initiated exactly $\tau$ steps after the initial stimulus (see \Cref{fig:counterexample}, center).
The operator corresponding to this update is $\mH \colon \vv \mapsto \vv + (\gamma \mP_\pi)^\tau (\mT_\pi \vv - \vv)$.
The fixed point of this operator is $\vv_\pi$ for any value of $\tau$ because $\mT_\pi \vv_\pi - \vv_\pi = 0$.

Notice that this is a rather benign form of non-recent credit assignment;
we are taking the simplest TD method and merely translating its impulse response along the time axis.
More complex forms of non-recent credit assignment would consist of a superposition of multiple such updates, and so this example provides insight into other methods.
Nevertheless, despite the simplicity of this method, we present a simple Markov reward process (MRP) that causes almost every value-function initialization to diverge away from $\vv_\pi$.

\begin{counterexample}
    \label{counterexample:pulse}
    Consider a 2-state MRP with reward $r(s,s') = 0$, $\forall~s,s' \in \{s_1,s_2\}$.
    Let ${p \in [0,1]}$ be the self-transition probability (see \Cref{fig:counterexample}, left) and let $\vv_0$ be the initial value function.
    If $\tau = 1$, $\gamma = 0.9$, and $p=0.4$, then the TD update in \Cref{eq:delayed_pulse} diverges whenever $\vv_0(s_1) \neq \vv_0(s_2)$.
    \looseness=-1
\end{counterexample}

We give specific values of $\tau$, $\gamma$, and $p$ for the sake of the counterexample;
however, it appears that divergence is inevitable for any $\tau > 0$ as ${\gamma \to 1}$ and ${p \to 0}$.
The divergent behavior of the method is visualized in \Cref{fig:counterexample} (right), where the arrows represent unit vectors pointing in the direction the expected update (i.e., $\mH \vv - \vv$).
Because the reward is zero for all transitions, $\vv_\pi$ is the origin (red star) regardless of $\gamma$ and $p$.
However, we see that every value-function initialization not on the blue line where $\vv_0(s_1) = \vv_0(s_2)$ progresses arbitrarily far away from the fixed point, $\vv_\pi$.

Why does violating the recency heuristic in this easy problem cause divergence?
The reason becomes more clear when we observe that
$\gamma^\tau \delta_{t+\tau} = \smash{\nstep{\tau+1}_t} - \smash{\nstep{\tau}_t}$.
Thus, an equivalent operator for \Cref{eq:delayed_pulse} is $\vv \mapsto \vv + \mT_\pi^{\tau+1} \vv - \mT_\pi^\tau \vv$, whose worst-case contraction modulus is $1 + \gamma^{\tau+1} + \gamma^\tau$ by the triangle inequality---greater than $1$.
Although this does not automatically mean the operator will diverge, it does suggest that divergence is possible, and we see one instance of it here.
It is important to note that this divergence is not due to sampling noise nor an uneven state distribution, as we are explicitly computing the expected result of the operator in both states.
Furthermore, the phenomenon is not unique to this particular algorithm or problem, but generally arises whenever the weak recency heuristic is violated too much.
We prove this formally in the next section.

\section{Only Convex Returns Satisfy the Weak Recency Heuristic}
\label{sec:convex_wrh}

Recall that convex returns are convex combinations of $n$-step returns:
either compound returns or $n$-step returns themselves.
In this section, we show this definition is logically equivalent to the weak recency heuristic;
\Cref{def:wrh} is satisfied if and only if a return estimate is convex (see \Cref{prop:wrh}).

To illuminate the role of the weak recency heuristic, we first justify the general return estimator in \Cref{eq:td_error_forward_view}.
In particular, we show that estimates of this form correspond to the largest set of linear operators suitable for TD learning.
This allows us to later analyze how the properties of these operators are affected by the choice of the weights, $(h_i)_{i=0}^\infty$, especially when these weights do not satisfy the recency heuristic.

To produce a TD method in the form of \Cref{eq:backup} that converges to $\vv_\pi$ under general conditions, the return estimate $\hat{G}_t$ must correspond to a maximum-norm contraction mapping, $\mH$, with its unique fixed point at $\vv_\pi$ (recall \Cref{sec:bg}).
In addition to these requirements, we want a \emph{sample-realizable} operator in order to create an implementable TD method:
one that can be constructed from any rewards or state values following time $t$.
To match existing TD methods, we assume that this operator is linear with respect to these quantities, giving us the following definition.

\begin{definition}
    \label{def:sample-real_op}
    A sample-realizable linear operator has the form
    ${\mH \vv \! = \! \sum_{i=0}^\infty a_i (\gamma \mP_\pi)^i \vr \! + \! b_i (\gamma \mP_\pi)^i \vv}$,
    where $(a_i)_{i=0}^\infty$ and $(b_i)_{i=0}^\infty$ are bounded sequences of real numbers.
\end{definition}

This definition covers all possible operators based on return estimates that can be constructed from a linear combination of sampled experiences: i.e.,
$\hat{G}_t = \sum_{i=0}^\infty a_i \gamma^i R_{t+i} + b_i \gamma^i V_{t+i}$.
However, the vast majority of these operators will not meet our convergence criteria.
In the following proposition, we reduce the space of operators by identifying only those whose fixed point is exactly $\vv_\pi$.

\begin{restatable}[]{prop}{propsamplerealop}
    \label{prop:sample-real_op}
    For every sample-realizable operator $\mH$ whose fixed point is $\vv_\pi$, there exists a sequence of real numbers $(h_i)_{i=0}^\infty$ such that
    \begin{equation}
        \label{eq:sample-real_errors}
        \mH \vv = \vv + \sum_{i=0}^\infty h_i (\gamma \mP_\pi)^i (\mT_\pi \vv - \vv)
        \,.
    \end{equation}
    If we let $c_n \defeq h_{n-1} - h_n$ for $n \geq 1$, then $\mH$ also has the equivalent form
    \begin{equation}
        \label{eq:sample-real_nsteps}
        \mH \vv
        = \left(1-\sum_{n=1}^\infty c_n\right) \! \vv + \sum_{n=1}^\infty c_n \mT_\pi^n \vv
        \,.
    \end{equation}
\end{restatable}

\begin{proof}
    See \Cref{subapp:prop_sample-real_op}.
\end{proof}

Notice that \Cref{eq:sample-real_errors} corresponds exactly to the sample estimate in \Cref{eq:td_error_forward_view} that we considered in \Cref{sec:recency_heuristic} when defining the weak recency heuristic.
We refer to these as \emph{linear} returns.
Hence, every linear return with $\vv_\pi$ as its fixed point is expressible as a weighted sum of either TD errors or $n$-step returns, without loss of generality.

We now have a generic operator that is both sample realizable and has the correct fixed point, but it is not necessarily a contraction mapping without any conditions on its weights, $(h_i)_{i=0}^\infty$\,.
\Cref{eq:sample-real_nsteps} expresses the operator in terms of the $n$-step Bellman operators, facilitating the analysis of its contraction properties.
Because $\mP_\pi$ is a stochastic matrix, we have
${\norm{\mP_\pi}_\infty = 1}$, which also implies that
$\norm{\mT_\pi^n \vv - \mT_\pi^n \vv'}_\infty \leq \gamma^n \norm{\vv - \vv'}_\infty$, for any $\vv, \vv' \in \mathbb{R}^\abs{\mathcal{S}}$.
Thus, by the triangle inequality,
\begin{equation}
    \label{eq:linear_contraction}
    \norm{\mH \vv - \mH \vv'}_\infty \leq \left(\abs{1 - \sum_{n=1}^\infty c_n} + \sum_{n=1}^\infty \abs{c_n} \gamma^n\right) \norm{\vv - \vv'}_\infty
    \,,
\end{equation}
and the contraction modulus is therefore
${\beta = \abs{1 - \sum_{n=1}^\infty c_n} + \sum_{n=1}^\infty \abs{c_n} \gamma^n}$.
The operator is a contraction mapping if and only if $\beta < 1$.

\begin{wrapfigure}{R}{0.5\textwidth}
    \centering
    \includegraphics[width=0.45\textwidth]{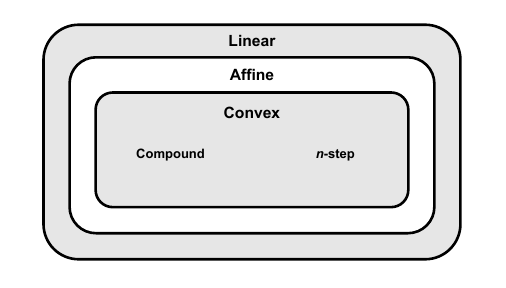}
    \caption{
        Hierarchical relationship between different return estimators.
        A return satisfies the weak recency heuristic if and only if it is a convex return:
        i.e., a compound or $n$-step return.
    }
    \label{fig:return_hierarchy}
\end{wrapfigure}

Notice that \Cref{eq:sample-real_nsteps} consists of two terms:
the original value function scaled by $1 - \sum_{n=1}^\infty c_n$, and a linear combination of $n$-step returns.
The first term can be eliminated without loss of generality by normalizing the sum of weights, i.e., by adding the constraint that $\sum_{n=1}^\infty c_n = 1$.
This is because the first term changes only the magnitude of the update, which can be absorbed into the step size, $\alpha_t$, in \Cref{eq:op_update}.
With this constraint in place, it follows that the weight of the first TD error is $h_0 = 1$ because of the telescoping series:
$h_0 = \sum_{n=1}^\infty h_{n-1} - h_n = \sum_{n=1}^\infty c_n = 1$.
The operator is now an affine combination of $n$-step Bellman operators, and so we refer to such return estimates as \emph{affine} returns.
Note that, since we have $h_0 = 0$ in \Cref{eq:delayed_pulse} when $\tau > 0$, the divergent return estimate in \Cref{counterexample:pulse} is \emph{not} an affine return, although it is linear.
Affine returns look identical to convex returns from \Cref{eq:convex_return}, but they are more general because they allow for negatively weighted $n$-step returns.
We depict the hierarchical relationship between linear, affine, convex, compound, and $n$-step returns in \Cref{fig:return_hierarchy}, and summarize their operators and corresponding sample estimates in \Cref{tab:estimators}.

This analysis provides a hint of why counterexamples like the one in \Cref{sec:counterexample} are possible;
negative weights increase the contraction modulus due to the absolute value in \Cref{eq:linear_contraction}.
It turns out that such negative weights coincide exactly with the time steps on which the weak recency heuristic is violated, and therefore only convex returns satisfy the heuristic, as we show in the next proposition.

\begin{restatable}[]{prop}{propwrh}
    \label{prop:wrh}
    An affine return satisfies the weak recency heuristic if and only if it is a convex return (i.e., a compound return or an $n$-step return).
\end{restatable}

\begin{proof}
    See \Cref{subapp:prop_wrh}.
\end{proof}

An immediate corollary of the above is that the weak recency heuristic is a sufficient condition for convergence, since both compound returns and $n$-step returns are already known to correspond to contraction mappings \citep[][Sec.~7.2]{watkins1989learning}.
This stems from the fact that a convex combination of $n$-step returns, each of which is contractive with modulus $\gamma^n$, must also be contractive: i.e.,
$\sum_{n=1}^\infty c_n \gamma^n \leq \gamma < 1$
for every choice of nonnegative weights that sum to one.
In this view, the weak recency heuristic can be seen as a convergence test for TD learning, and explains some of its utility in computational RL:
divergence is impossible under this heuristic.

On the other hand, violating the weak recency heuristic increases the contraction modulus of the return estimator, with divergence possible if the violation becomes too extreme (e.g., \Cref{counterexample:pulse}).
This is because any time an $n$-step return has a negative weight, another $n$-step return must have a larger positive weight to counterbalance it and ensure the weights sum to $1$ overall.
This necessarily increases the contraction modulus in \Cref{eq:linear_contraction} due to the absolute value, underscoring yet another benefit of the weak recency heuristic.
A convex return is not only guaranteed to converge regardless of its weights, but also has a faster contraction than a nonconvex (affine) return constructed from the same $n$-step returns.

\begin{table}[t]
    \renewcommand{\arraystretch}{3}
    \footnotesize
    \centering
    \begin{tabular}{llll}
        \toprule
         Name & Operator & Sample Estimate & Conditions \\
         \midrule
         Linear
            & $\Big(1 \! - \!\! \sum\limits_{n=1}^\infty c_n\Big) \vv + \sum\limits_{n=1}^\infty c_n \mT_\pi^n \vv$
            & $\Big(1 \! - \!\! \sum\limits_{n=1}^\infty c_n\Big) V_t + \sum\limits_{n=1}^\infty c_n G^n_t$
            & None \\
         Affine
            & $\sum\limits_{n=1}^\infty c_n \mT_\pi^n \vv$
            & $\sum\limits_{n=1}^\infty c_n G^{n}_t$
            & $\sum\limits_{n=1}^\infty c_n = 1$ and $\sum\limits_{n=1}^\infty \abs{c_n} \gamma^n < 1$ \\
         Convex
            & $\sum\limits_{n=1}^\infty c_n \mT_\pi^n \vv$
            & $\sum\limits_{n=1}^\infty c_n G^{n}_t$
            & Affine and $c_n \geq 0$, $\forall~n \geq 1$ \\
         Compound
            & $\sum\limits_{n=1}^\infty c_n \mT_\pi^n \vv$
            & $\sum\limits_{n=1}^\infty c_n G^{n}_t$
            & Convex and $\exists~c_i, c_j > 0$ \\
         $n$-step
            & $\mT_\pi^n v$
            & $G^n_t$
            & $n \geq 1$ \\
         \bottomrule
    \end{tabular}
    \caption{Summary of operators and sample estimates for the return estimators in \Cref{fig:return_hierarchy}.}
    \label{tab:estimators}
\end{table}

\section{Are Monotonically Decreasing Weights Necessary?}

So far, we have focused on the weak recency heuristic:
when the eligibility weights are nonincreasing.
However, as we discussed in \Cref{sec:recency_heuristic}, the connotation of the recency heuristic is often that of strictly decreasing TD-error weights, i.e., the strong recency heuristic (\Cref{def:srh}).
This is why, for example, $\lambda$-returns are more strongly associated with a recency heuristic than $n$-step returns are.
Does this distinction between weak and strong recency heuristics matter in practice?
In this section, we conduct experiments indicating that the answer is yes, but in a surprising way;
the smoothness of the weights do not appear to be significant, but the strong recency heuristic does imply that the return estimate consists of infinitely many $n$-step returns, which empirically improves credit assignment.
\looseness=-1

\begin{wrapfigure}{R}{0.72\textwidth}
    \centering
    \begin{minipage}[b]{0.34\textwidth}
        \centering
        \includegraphics[width=\textwidth]{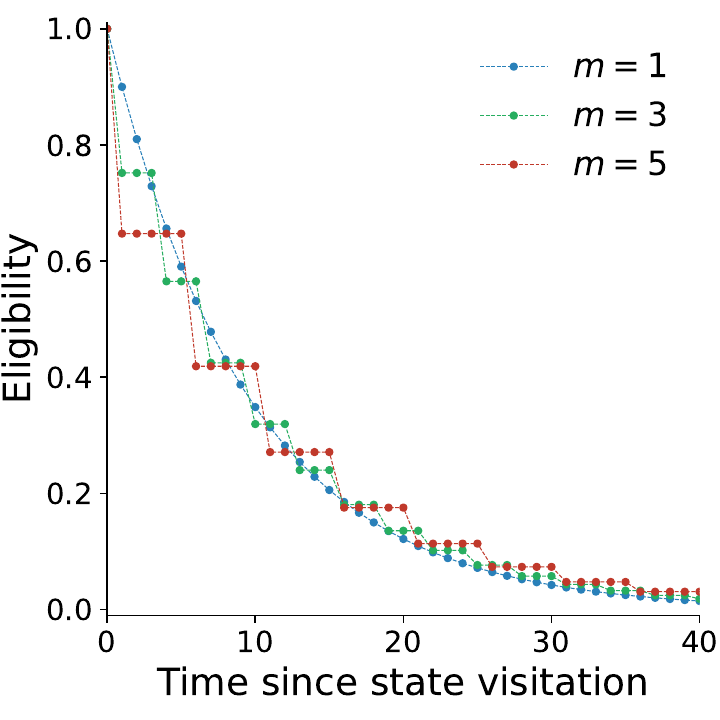}
        \caption{Impulse responses of $\lambda$-returns with varying degrees of sparsity.}
        \label{fig:td-error_weights_sparse}
    \end{minipage}
    \hfill
    \begin{minipage}[b]{0.34\textwidth}
        \centering
        \includegraphics[width=\textwidth]{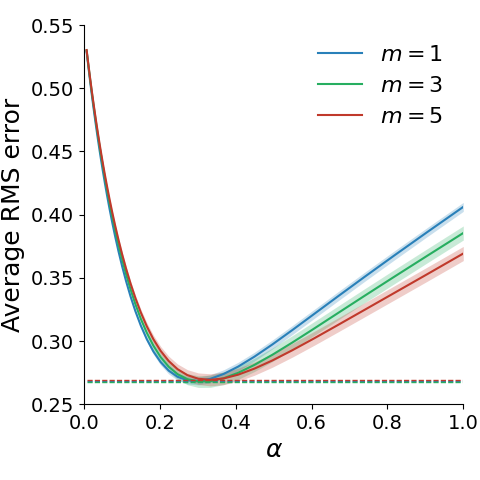}
        \caption{Random-walk performance of $\lambda$-returns with varying degrees of sparsity.}
        \label{fig:rw19_sparse}
    \end{minipage}
    \hfill
\end{wrapfigure}

To test the question of whether the smoothness of the TD-error weights matters, we introduce the \emph{sparse} $\lambda$-return, defined as
\begin{equation}
    \label{eq:sparse_lambda}
    G^{\lambda,m}_t
    \defeq \sum_{i=0}^\infty \gamma^i \lambda^{\lfloor \frac{i+m-1}{m} \rfloor} \delta_{t+i}
    = (1-\lambda) \sum_{k=1}^\infty \lambda^{k-1} \nstep{m(k-1)+1}_t
    \,,
\end{equation}
where $m \geq 1$.
The contraction modulus of this return is $\beta = \gamma (1-\lambda) \mathbin{/} (1 - \gamma^m \lambda)$.
When $m=1$, we simply recover the standard exponential decay of the $\lambda$-return from \Cref{eq:lambda-return_nsteps}.
However, for $m > 1$, the TD-error weights no longer satisfy the strong recency heuristic as they become more stepwise (see \Cref{fig:td-error_weights_sparse}).
This implies that every $m-1$ out of $m$ $n$-step returns have zero weight.
For example, setting $m=2$ generates the TD-error weight sequence
$(1,\ \lambda,\ \lambda,\ \lambda^2,\ \lambda^2,\ \dots)$, 
which produces an exponential average of the odd $n$-step returns:
$(\smash{\nstep{1}_t}\mkern-12mu,\ \smash{\nstep{3}_t}\mkern-12mu,\ \smash{\nstep{5}_t}\mkern-12mu,\ \smash{\nstep{7}_t}\mkern-12mu,\ \smash{\nstep{9}_t}\mkern-12mu,\ \dots)$.
The reason we choose this form is because it isolates the effects of the two recency heuristics by keeping the type of weighted average consistent (i.e., exponential).
If monotonicity is beneficial to learning, then we would expect to observe a performance degradation for sparse $\lambda$-returns ($m > 1$) compared to dense ($m=1$).

\begin{figure}[b]
    \centering
    \includegraphics[width=\textwidth]{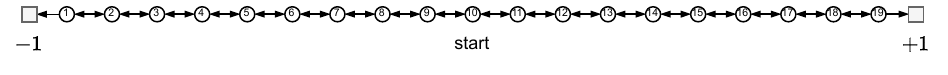}
    \vspace{-0.35in}  
    \caption{The 19-state random walk \citep[][Sec.~12.1]{sutton2018reinforcement}.}
    \label{fig:rw19}
\end{figure}

Our experiment setup is a discounted variation ($\gamma=0.99$) of the 19-state random walk from \citet[][Sec.~12.1]{sutton2018reinforcement}.
In this environment, each episode starts with the agent in the center of a linear chain of 19 connected states (see \Cref{fig:rw19}).
The agent can move either left or right, and its behavior is fixed such that it randomly chooses either action with equal probability.
Reaching either end of the chain terminates the episode and yields a reward:
$-1$ for the left or $+1$ for the right.
\looseness=-1

We test three different degrees of sparsity for the $\lambda$-returns, adjusting $\lambda$ for each return to maintain the same contraction modulus in all cases:
$(\lambda,m) \in \{(0.9,1), (0.75,3), (0.65,5)\}$.
The agents are trained for 10 episodes by applying offline value backups of the form \Cref{eq:backup} to every experience at the end of each episode.
In \Cref{fig:rw19_sparse}, we plot the root-mean-square (RMS) error, $\norm{\vv - \vv_\pi}_2$, averaged over the 10 episodes versus the step size, $\alpha$, for each return.
The final results are averaged over 400 trials with 95\% confidence intervals indicated by shaded regions.
Code is available online.\footnote{
    \url{https://github.com/brett-daley/recency-heuristic}
}

Because the three returns all have the same contraction modulus (i.e., expected convergence rate), their performance is nearly the same for small values of $\alpha$ which are able to average out the noise in the updates.
Likewise, the returns share the same lowest error, as indicated by the dashed horizontal lines in \Cref{fig:rw19_sparse}.
However, as $\alpha$ gets larger, their performance begins to separate, achieving lower average error as the sparsity of the $\lambda$-return increases.
Thus, even though the eligibility curves become more step-like as the sparsity is increased and they violate the strong recency heuristic, the overall performance of the return improves.
This demonstrates that the monotonicity of the eligibility curves does not directly factor into the performance of the return estimators.

The main reason for the sparse $\lambda$-return's improvement appears to be that its eligibility initially decays faster than that of the dense $\lambda$-return, but then slower as time goes on (see \Cref{fig:td-error_weights_sparse}).
This gives the eligibility curve a long-tailed characteristic which, in turn, propagates credit back in time more quickly.
In fact, every return that satisfies the strong recency heuristic must have a similar characteristic, because \Cref{def:srh} implies that $c_n = h_{n-1} - h_n > 0$ for all $n \geq 1$, and thus \Cref{eq:convex_return} must correspond to a positively weighted average of infinitely many $n$-step returns.
Although this property is not unique to the strong recency heuristic (e.g., the sparse $\lambda$-return has it but does not satisfy \Cref{def:srh}), it does suggest a practical significance for this heuristic:
it implies a longer horizon for credit assignment.

However, any benefit of a longer credit-assignment horizon is contingent on controlling the variance of the return.
Fortunately, as we show in the following proposition, a long-tailed eligibility curve does not increase the worst-case variance when the contraction modulus is held constant.

\begin{restatable}[]{prop}{propvariance}
    \label{prop:variance}
    Let $\kappa_t \defeq \max_{i,j \geq 0} \Cov[\delta_{t+i},\delta_{t+j} \mid S_t]$.
    The worst-case conditional variance of any convex return $G^\vc_t$ with contraction modulus $\beta$ has the bound
    \begin{equation}
        \label{eq:variance_bound}
        \Var[G^\vc_t \mid S_t]
        \leq \left( \frac{1-\beta}{1-\gamma} \right)^{\!2} \kappa_t
        \,.
    \end{equation}
\end{restatable}

\begin{proof}
    See \Cref{subapp:prop_variance}.
\end{proof}

This bound is rather loose, but it is general.
\Cref{eq:variance_bound} implies that averages of $n$-step returns always have finite variance, even as the $n$-step returns become arbitrarily long.
Furthermore, this upper bound depends only on the contraction modulus of the return itself and not the chosen weights for the average.
Since the contraction modulus is proportional to the worst-case bias of the return by \Cref{eq:linear_contraction}, we see that both the worst-case bias and worst-case variance of the $\lambda$-returns in our previous experiment remain the same regardless of sparsity.
Thus, compound returns with a long-tailed eligibility curve are able to assign credit more quickly without negatively impacting the bias-variance trade-off\footnote{
    In fact, it is likely such long-tailed returns have a \emph{positive} impact on the bias-variance trade-off by reducing variance, under an additional assumption that the TD-error variances are roughly uniform \citep[see][Sec.~6]{daley2024averaging}.
    \looseness=-1
}
(at least, in a worst-case sense).

\begin{wrapfigure}{R}{0.72\textwidth}
    \centering
    \begin{minipage}[b]{0.34\textwidth}
        \centering
        \includegraphics[width=\textwidth]{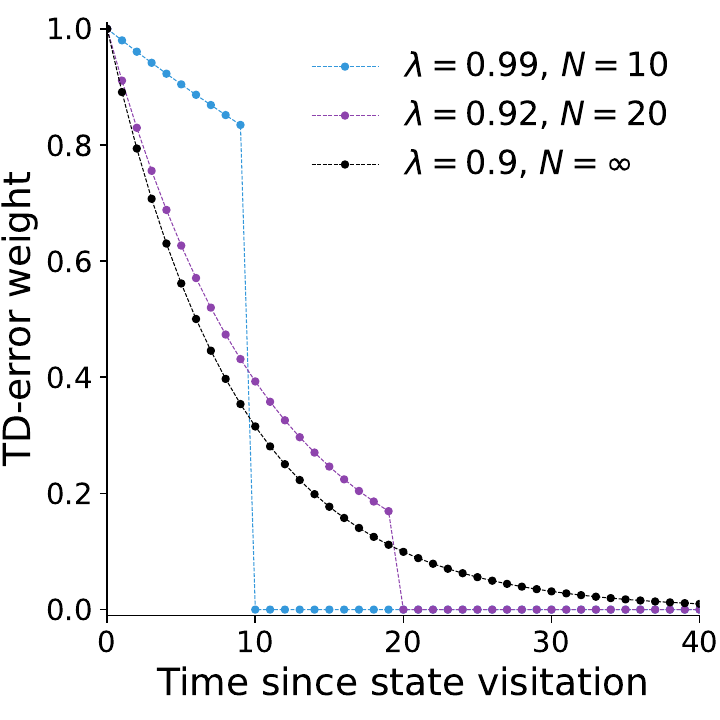}
        \caption{Eligibility curves of $\lambda$-returns with varying degrees of truncation.}
        \label{fig:td-error_weights_trunc}
    \end{minipage}
    \hfill
    \begin{minipage}[b]{0.34\textwidth}
        \centering
        \includegraphics[width=\textwidth]{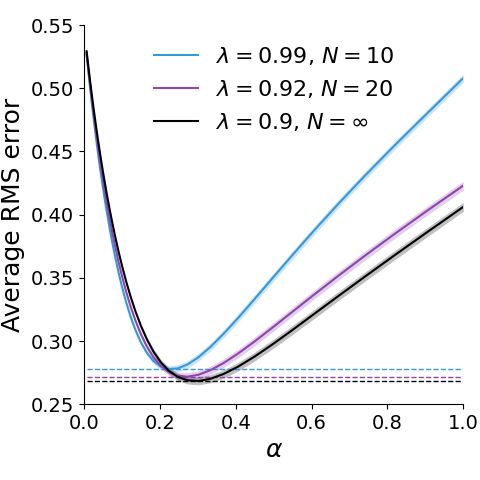}
        \caption{Random-walk performance of $\lambda$-returns with varying degrees of truncation.}
        \label{fig:rw19_trunc}
    \end{minipage}
    \hfill
\end{wrapfigure}

To test the effect of a longer credit-assignment horizon under a controlled contraction modulus, we repeat the previous random-walk experiment but with \emph{truncated} $\lambda$-returns:
$\smash{G^\lambda_{t:t+N}}\ \smash{\defeq}\ \smash{V_t + \sum_{i=0}^{N-1} (\gamma \lambda)^i \delta_{t+i}} = \smash{(1-\lambda) \sum_{n=1}^{N-1} \lambda^{n-1} \smash{\nstep{n}_t} + \lambda^{N-1} \nstep{N}_t}$,
where $N \geq 1$ is the truncation length.
The contraction modulus of this return is
$\beta = \big((1-\gamma)(\gamma\lambda)^N + \gamma(1-\lambda)\big) \mathbin{/} (1-\gamma\lambda)$.
The eligibility curve for this return is a monotonically decreasing function, up until time $N$ when it abruptly falls to zero (see \Cref{fig:td-error_weights_trunc}).
As $N \to \infty$, we recover the true $\lambda$-return, \Cref{eq:lambda-return_nsteps}.
We test three variants of this return:
$(\lambda,N) \in \{(0.99,10), (0.93,20), (0.9,\infty)\}$.
As before, all of these values are chosen to produce approximately the same contraction modulus.
We plot the average RMS error in \Cref{fig:rw19_trunc}, again averaged over 400 trials with 95\% confidence intervals shaded.
The performance is roughly identical when $\alpha$ is small, since the same contraction modulus guarantees the same expected performance.
However, as $\alpha$ gets larger, the truncated returns perform poorly compared to the full $\lambda$-return.
This suggests that the performance of the returns is strongly tied to longer $n$-step returns in the average, but only when the contraction moduli are equalized.
This also supports our earlier hypothesis that the results observed with the sparse $\lambda$-returns in \Cref{fig:rw19_sparse} are due to their long-tail eligibility curves and not some other property such as monotonicity.

To summarize, satisfying the strong recency heuristic creates a compound return consisting of infinitely many $n$-step returns---a long-tailed eligibility curve.
This improves the effective window of credit assignment without exacerbating variance (in a conservative sense), as long as the contraction modulus is held constant.
However, this property is not unique to the strong recency heuristic;
for instance, sparse $\lambda$-returns violate this heuristic, but are still averages of infinitely many $n$-step returns, and outperform dense $\lambda$-returns in \Cref{fig:rw19_sparse}.
These insights help explain why smooth averages like the $\lambda$-return are often effective in practice, even if not strictly necessary for good performance.

\section{Off-Policy Learning and Other Extensions}
\label{sec:extensions}

The weak recency heuristic is closely tied to an open problem on the convergence of off-policy eligibility traces \citep[][Sec.~5.3]{daley2023trajectory}.
Off-policy learning occurs whenever the agent's policy for action selection, $b$, differs from the policy for return estimation, $\pi$.
Let $\smash{\rho_{t+i} \defeq \pi(A_{t+i}|S_{t+i}) \mathbin{/} b(A_{t+i}|S_{t+i})}$ be the importance-sampling ratio.
\citet{daley2023trajectory} proved that satisfying
$h_i \rho_{t+i+1} \geq h_{i+1} \geq 0$, $\forall~i \geq 0$,
is sufficient for the off-policy update analogous to \Cref{eq:td_error_forward_view} to converge to $\vv_\pi$, where the TD-error weights can generally be \emph{trajectory aware} (i.e., dependent on past state-action pairs).
The open problem is to determine whether this condition is necessary as well.
\looseness=-1

Notably, the condition is exactly the off-policy generalization of the weak recency heuristic (\Cref{def:wrh}), since $\rho_{t+i} = 1$ when $\pi = b$.
Based on the analysis in \Cref{sec:convex_wrh}, we know that it is sometimes possible to violate this heuristic and still converge, and so \emph{the condition is sufficient but not necessary}.
We provide more details in \Cref{app:extensions}, where we also extend our theory to state-dependent eligibilities \citep[e.g.,][]{yu2012least,white2016greedy} and function approximation \citep{tsitsiklis1997analysis}.
These results show that the possibility of divergence like in \Cref{counterexample:pulse} is a general phenomenon of TD learning when not utilizing a recency heuristic.

\hfill

\section{Conclusion}

Although non-recent credit assignment
should theoretically be possible and useful in certain learning environments, it does not seem readily compatible with our current formulation of TD learning.
In particular, violating the recency heuristic manifests as negative weights on some of the $n$-step components of the return target.
These negative weights appear to counteract learning by increasing the contraction modulus, without offering a clear benefit to learning, and potentially culminating in divergence as demonstrated by \Cref{counterexample:pulse}.
The fact that divergence is possible in such a favorable setting---an on-policy, tabular MRP with fully observable states---points to the severity of this issue.
Indeed, as we discussed in \Cref{sec:extensions}, this issue persists in more challenging settings including off-policy learning and function approximation.
Successfully implementing new forms of credit assignment that do not strictly follow the recency heuristic will likely require rethinking how we formulate the reinforcement signal in computational RL.
Our theory will provide a good starting point for algorithmic development in this direction.

Another major finding is that the recency heuristic is not merely a simple protocol for addressing the temporal credit-assignment problem, but also has intrinsic importance for learning value functions.
The existence of diverging counterexamples illuminates the critical role of nonincreasing weights on the TD errors---the weak recency heuristic.
The logical equivalence between this heuristic and the return estimate's ability to be expressed as a convex combination of $n$-step returns unifies two fundamental yet seemingly disparate ideas in RL.
More specifically, convex returns were the most general return estimates for TD learning identified before our work, and so it is surprising to find they coincide exactly with another foundational concept in RL:
the recency heuristic.
This appears to be a novel, unifying perspective between the forward and backward views of TD learning with arbitrary return estimates.
In the off-policy setting, the weak recency heuristic is equivalent to the convergence condition for eligibility traces discovered by \citet[][]{daley2023trajectory}, providing more evidence for its importance in learning value functions.

Finally, our results help to further explain the strong empirical performance and continued popularity of TD($\lambda$), along with its many variants, for nearly four decades.
Our experiments suggest that the smoothness of TD($\lambda$)'s exponential decay is not directly responsible for this success;
rather, all compound returns (including $\lambda$-returns) that average an infinite number of $n$-step returns are able to distribute credit over a longer period without exacerbating the maximum bias or variance.
These results confirm the intuition that ``the fading strategy [of TD($\lambda$)] is often the best [versus $n$-step TD methods]'' \citep[][p.~304]{sutton2018reinforcement}, though non-exponential fading strategies are also viable.

%% file: paper_appendix.tex
\appendix
\clearpage

\section{Proofs}
\label{app:proofs}

This section contains the omitted proofs of all propositions in the paper.

\subsection{Proof of \Cref{prop:sample-real_op}}
\label{subapp:prop_sample-real_op}

\propsamplerealop*

\begin{proof}
It is given that $\mH$ is sample realizable.
Without loss of generality, we consider an alternative parameterization of \Cref{def:sample-real_op} that spans the same space of linear operators.
There exist sequences of real numbers $(x_i)_{i=0}^\infty$ and $(y_i)_{i=0}^\infty$ such that
\begin{align*}
    \mH \vv
    &= \vv + \sum_{i=0}^\infty (\gamma \mP_\pi)^i \Big[ x_i (\vr + \gamma \mP_\pi \vv) - y_i \vv \Big] \\
    &= \vv + \sum_{i=0}^\infty (\gamma \mP_\pi)^i \Big[ x_i (\vr + \gamma \mP_\pi \vv - \vv) + (x_i - y_i) \vv \Big] \\
    &= \vv + \sum_{i=0}^\infty (\gamma \mP_\pi)^i \Big[ x_i (\mT_\pi \vv - \vv) + (x_i - y_i) \vv \Big] \\
    \label{eq:alt-sample-real}
    &= \vv + \sum_{i=0}^\infty x_i (\gamma \mP_\pi)^i (\mT_\pi \vv - \vv) + \sum_{i=0}^\infty (x_i - y_i) (\gamma \mP_\pi)^i \vv
    \,.
\end{align*}
Because $\mT_\pi \vv_\pi = \vv_\pi$, it follows that
$\mH \vv_\pi = \vv_\pi + \sum_{i=0}^\infty (x_i - y_i) (\gamma \mP_\pi)^i \vv_\pi$\,.
To ensure that $\vv_\pi$ is the fixed point of $\mH$ (i.e., that $\mH \vv_\pi = \vv_\pi$), we must make the remaining sum zero.
However, this happens only when $x_i = y_i$, $\forall~i \geq 0$\,.
Thus, we substitute $h_i = x_i$ and $h_i = y_i$ to get \Cref{eq:sample-real_errors}.
\looseness=-1

To derive \Cref{eq:sample-real_nsteps}, we apply the fact that $h_i = \sum_{n=i+1}^\infty c_n$ due to the telescoping series.
We complete the proof by rewriting \Cref{eq:sample-real_errors} as
\begin{align*}
    \mH \vv
    &= \vv + \sum_{i=0}^\infty \left( \sum_{n=i+1}^\infty c_n \right) (\gamma \mP_\pi)^i (\mT_\pi \vv - \vv) \\
    &= \vv + \sum_{n=1}^\infty c_n \sum_{i=0}^{n-1} (\gamma \mP_\pi)^i (\mT_\pi \vv - \vv) \\
    &= \vv + \sum_{n=1}^\infty c_n (\mT_\pi^n \vv - \vv) \\
    &= \left(1-\sum_{n=1}^\infty c_n\right) \vv + \sum_{n=1}^\infty c_n \mT_\pi^n \vv
    \,.
\end{align*}
The second equality interchanged the sums using the rule
$\sum_{i=0}^\infty \sum_{n=i+1}^\infty = \sum_{n=1}^\infty \sum_{i=0}^{n-1}$.
The third equality followed from the $n$-step Bellman operator expansion:
$\mT_\pi^n \vv = \vv + \sum_{i=0}^{n-1} (\gamma \mP_\pi)^i (\mT_\pi \vv - \vv)$.
\end{proof}

\subsection{Proof of \Cref{prop:wrh}}
\label{subapp:prop_wrh}

\propwrh*

\begin{proof}
Recall that $c_n = h_{n-1} - h_n$.
Therefore, the affine operator from \Cref{eq:sample-real_nsteps} is equal to
\begin{equation}
    \label{eq:convex_combo}
    \mH \vv
    = \sum_{n=1}^\infty \mathop{(h_{n-1} - h_n)} \mT_\pi^n \vv
    \,.
\end{equation}
If the weak recency heuristic (\Cref{def:wrh}) holds, then we have ${h_{n-1} \geq h_n} \implies {h_{n-1} - h_n \geq 0}$, for all $n \geq 1$.
Thus, \Cref{eq:convex_combo} is a convex combination of $n$-step returns, because we have ${\sum_{n=1}^\infty h_{n-1} - h_n} = \sum_{n=1}^\infty c_n = 1$ for an affine return.

To complete the proof, we also show the contrapositive.
Consider an affine return that is not a convex combination of $n$-step returns.
Consequently, it must have at least one negatively weighted $n$-step return:
there exists some $k \geq 1$ such that $c_k < 0$.
However, this implies that ${h_{k-1} - h_k < 0}$, and therefore ${h_{k-1} < h_k}$, so the weak recency heuristic is violated.
We conclude that an affine return satisfies the weak recency heuristic if and only if it is a convex return.
\end{proof}

\subsection{Proof of \Cref{prop:variance}}
\label{subapp:prop_variance}

\propvariance*

\begin{proof}
First, note that
$\Var[\hat{G}_t \mid S_t] = \Var[\hat{G}_t - V_t \mid S_t]$
for any return estimate, $\hat{G}_t$, since $V_t$ is deterministic given state $S_t$.
This allows us to derive an upper bound on the covariance between two $n$-step returns with lengths $n_1$ and $n_2$ using \Cref{eq:nstep-return_errors}:
\begin{align*}
    \Cov[\nstep{n_1}_t,\nstep{n_2}_t \mid S_t]
    &= \Cov\Bigg[\sum_{i=0}^{n_1-1} \gamma^i \delta_{t+i}, \sum_{j=0}^{n_2-1} \gamma^j \delta_{t+j} \Biggm| S_t\Bigg] \\
    &= \sum_{i=0}^{n_1-1} \sum_{j=0}^{n_2-1} \gamma^{i+j} \Cov[\delta_{t+i}, \delta_{t+j} \mid S_t] \\
    &\leq \sum_{i=0}^{n_1-1} \sum_{j=0}^{n_2-1} \gamma^{i+j} \kappa_t \\
    &= \gammafunc{n_1} \gammafunc{n_2} \kappa_t
    \,,
\end{align*}
where $\gammafunc{n} \defeq (1 - \gamma^n) \mathbin{/} (1-\gamma)$ is the $n$-th partial sum of the geometric series.
Because $\sum_{n=1}^\infty c_n = 1$ and $\beta = \sum_{n=1}^\infty c_n \gamma^n$ for a convex return, we also have
\begin{equation*}
    \sum_{n=1}^\infty c_n \gammafunc{n}
    = \sum_{n=1}^\infty c_n \left(\frac{1-\gamma^n}{1-\gamma}\right)
    = \frac{1 - \sum_{n=1}^\infty c_n \gamma^n}{1-\gamma}
    = \frac{1 - \beta}{1-\gamma}
    \,.
\end{equation*}
Therefore, we derive the following upper bound on the variance of a convex return:
\begin{align*}
    \Var[G^\vc_t \mid S_t]
    &= \sum_{i=1}^\infty \sum_{j=1}^\infty \Cov[c_i \nstep{i}_t, c_j \nstep{j}_t \mid S_t] \\*
    &= \sum_{i=1}^\infty \sum_{j=1}^\infty c_i c_j \Cov[\nstep{i}_t, \nstep{j}_t \mid S_t] \\*
    &\leq \sum_{i=1}^\infty \sum_{j=1}^\infty c_i c_j \gammafunc{i} \gammafunc{j} \kappa_t \\*
    &= \left(\frac{1-\beta}{1-\gamma}\right)^{\!2} \kappa_t
    \,,
\end{align*}
which completes the proof.
\end{proof}

\section{Extensions}
\label{app:extensions}

This section contains extensions of our theory to off-policy learning, state- or trajectory-dependent eligibility traces, and function approximation.

\subsection{Function Approximation}

Our results easily generalize to the case where the value function is approximated by a linear parametric function:
$V_t = \vx_t\tran \vw_t$, where $\vw_t \in \mathbb{R}^d$ is the value-function weights, and $\vx_t \in \mathbb{R}^d$ is a feature vector corresponding to state $S_t$.
Because $\frac{\partial}{\partial \vw} V_t\bigr|_{\vw=\vw_t} = \vx_t$, the semi-gradient TD update becomes
\begin{equation*}
    \vw_{t+1} = \vw_t + \alpha_t \mathop{\big( \hat{G}_t - V_t \big)} \vx_t
    \,.
\end{equation*}
Let $\mX \in \mathbb{R}^{\abs{\mathcal{S}} \times d}$ be the matrix whose rows correspond to the feature vectors for every state in $\mathcal{S}$.
Because $\mH \vv$ generally cannot be represented exactly by the function approximator, the estimate $\hat{G}_t$ corresponds to a composite linear operator $\mPi \mH$, where $\mPi$ is a projection operator onto the set $\{\mX \vw \mid \vw \in \mathbb{R}^d\}$ under the state weighting induced by the MDP's stationary distribution \citep{tsitsiklis1997analysis}.
Furthermore, $\mPi$ is nonexpansive, linear, and independent of $\vw_t$ \citep[][proof of Lemma~6]{tsitsiklis1997analysis};
hence, if $\mH$ is a contraction mapping, then so is $\mPi \mH$ with the same maximum contraction modulus.
This implies that violating the weak recency heuristic too much can still increase the contraction modulus and cause divergence, just like in \Cref{counterexample:pulse}.

In the case of \emph{nonlinear} function approximation, the existence of counterexamples is certain, as even TD(0) diverges for at least one function \citep[][Fig.~1]{tsitsiklis1997analysis}.

\subsection{State-Dependent Eligibility Traces}

The general return estimate considered by our work, \Cref{eq:td_error_forward_view}, determines the eligibility weights solely based on the elapsed time since the initial state.
Additionally, we can have weights that depend on the actual states experienced on each time step \citep[e.g.,][]{yu2012least,white2016greedy}.
A return estimate in this case has the form
\begin{equation}
    \label{eq:state_lambda}
    \hat{G}_t
    = V_t + \sum_{i=0}^\infty \mathop{h_i(S_{t+i})} \gamma^i \delta_{t+i}
    \,,
\end{equation}
where $h_i \colon \mathcal{S} \to \mathbb{R}$ is now a weighting function over the state space.
This estimate satisfies the weak recency heuristic if
\begin{equation*}
    h_i(s) \geq h_{i+1}(s') \geq 0\,,\enskip \forall~i \geq 0\,,\enskip
    \forall~s,s' \in \mathcal{S}
    \,.
\end{equation*}
The operator corresponding to \Cref{eq:state_lambda} is
$(\mH \vv)(s) = \mathbb{E}_\pi[\hat{G}_t \mid S_t=s]$, i.e., a convex combination of the estimates in \Cref{eq:state_lambda}.
Therefore, it too satisfies the weak recency heuristic, except that the weight at each time step is an average of random variables and cannot be explicitly written without additional information about the MDP.
Other than this minor difference, we see that the results for state-based eligibility curves are analogous to the strictly time-based eligibility curves discussed in our paper.
\looseness=-1

\subsection{Off-Policy Learning and Trajectory-Aware Eligibility Traces}

A further generalization of the state-dependent eligibility traces discussed in the previous section is trajectory-aware eligibility traces \citep{daley2023trajectory}.
These have been studied in the context of off-policy learning with action values, where the agent estimates the action-value function
$\smash{q_\pi(s,a) \defeq \mathbb{E}[G_t \mid (S_t,A_t)=(s,a)]}$.
Additionally, it is assumed that the agent samples actions from a behavior policy, $b$, that differs from the target policy, $\pi$.
The off-policy bias resulting from the mismatch between behavior and target distributions must be corrected to converge to $\vq_\pi$.

Let
$\smash{\mathcal{F}_{t:t+i} \defeq (S_{t+j},A_{t+j})_{j=0}^i}$
be the partial history of the MDP from time $t$ to $t+i$.
Additionally, let
${\delta^\pi_t \defeq R_t + \gamma \bar{V}_{t+1} - q(S_t,A_t)}$
denote the mean TD error using action values, where
${\bar{V}_t \defeq \sum_{a' \in \mathcal{A}} \mathop{\pi(a'|S_t)} q(S_t,a')}$.
A trajectory-aware return estimate has the form
\begin{equation*}
    \hat{G}_t
    = V_t + \sum_{i=0}^\infty \mathop{h_i(\mathcal{F}_{t:t+i})} \gamma^i \delta_{t+i}
    \,,
\end{equation*}
where $h_i \colon (\mathcal{S} \times \mathcal{A})^i \to \mathbb{R}$ is a weighting function over partial histories.
The corresponding operator is
$(\mH \vq)(s,a) = \mathbb{E}_\mu[\hat{G}_t \mid (S_t,A_t)=(s,a)]$.
For the operator to converge to $\vq_\pi$, it is sufficient to satisfy the following condition \citep[][Theorem~5.2]{daley2023trajectory}:
\begin{equation}
    \label{eq:trajectory-aware_condition}
    \mathop{h_i(\mathcal{F}_{t:t+i})} \rho_{t+i+1} \geq h_{i+1}(\mathcal{F}_{t:t+i+1}) \geq 0\,,\enskip \forall~i \geq 0\,,\enskip
    \forall~t \geq 0
    \,,
\end{equation}
where $\smash{\rho_{t+i} \defeq \pi(A_{t+i}|S_{t+i}) \mathbin{/} b(A_{t+i}|S_{t+i})}$ is the importance-sampling ratio.
An open problem is whether this condition is necessary in addition to being sufficient \citep[][Sec.~5.3]{daley2023trajectory}.
Rather interestingly, this condition is the off-policy analog of the weak recency heuristic, since $\mathbb{E}_\mu[\rho_{t+i+1} \mid (S_t,A_t)] = 1$ and therefore the inequality equates to \Cref{def:wrh} in expectation.
Based on our analysis in \Cref{sec:convex_wrh}, the heuristic can be slightly violated without increasing the contraction modulus above $1$, still allowing the operator to sometimes converge to $\vq_\pi$.
We thus settle the open problem in the negative:
the condition in \Cref{eq:trajectory-aware_condition} is \emph{sufficient but not necessary} for the operator to converge to its fixed point.

%% file: main.bbl
\begin{thebibliography}{31}
\providecommand{\natexlab}[1]{#1}
\providecommand{\url}[1]{\texttt{#1}}
\expandafter\ifx\csname urlstyle\endcsname\relax
  \providecommand{\doi}[1]{doi: #1}\else
  \providecommand{\doi}{doi: \begingroup \urlstyle{rm}\Url}\fi

\bibitem[Barto et~al.(1983)Barto, Sutton, and Anderson]{barto1983neuronlike}
Andrew~G. Barto, Richard~S. Sutton, and Charles~W. Anderson.
\newblock Neuronlike adaptive elements that can solve difficult learning control problems.
\newblock \emph{IEEE Transactions on Systems, Man, and Cybernetics}, 13\penalty0 (5):\penalty0 834--846, 1983.

\bibitem[Bellman(1957)]{bellman1957dynamic}
Richard Bellman.
\newblock \emph{Dynamic Programming}.
\newblock Princeton University Press, 1957.

\bibitem[Bertsekas \& Tsitsiklis(1996)Bertsekas and Tsitsiklis]{bertsekas1996neuro}
Dimitri~P. Bertsekas and John~N. Tsitsiklis.
\newblock \emph{Neuro-Dynamic Programming}.
\newblock Athena Scientific, 1996.

\bibitem[Cichosz(1995)]{cichosz1995truncating}
Pawel Cichosz.
\newblock Truncating temporal differences: On the efficient implementation of {TD($\lambda$)} for reinforcement learning.
\newblock \emph{Journal of Artificial Intelligence Research}, 2:\penalty0 287--318, 1995.

\bibitem[Daley \& Amato(2019)Daley and Amato]{daley2019reconciling}
Brett Daley and Christopher Amato.
\newblock Reconciling $\lambda$-returns with experience replay.
\newblock In \emph{Neural Information Processing Systems (NeurIPS)}, 2019.

\bibitem[Daley et~al.(2023)Daley, White, Amato, and Machado]{daley2023trajectory}
Brett Daley, Martha White, Christopher Amato, and Marlos~C. Machado.
\newblock Trajectory-aware eligibility traces for off-policy reinforcement learning.
\newblock In \emph{International Conference on Machine Learning (ICML)}, 2023.

\bibitem[Daley et~al.(2024)Daley, White, and Machado]{daley2024averaging}
Brett Daley, Martha White, and Marlos~C. Machado.
\newblock Averaging $n$-step returns reduce variance in reinforcement learning.
\newblock In \emph{International Conference on Machine Learning (ICML)}, 2024.

\bibitem[Gupta et~al.(2023)Gupta, Jordan, Chaudhari, Liu, Thomas, and da~Silva]{gupta2023past}
Dhawal Gupta, Scott~M Jordan, Shreyas Chaudhari, Bo~Liu, Philip~S. Thomas, and Bruno~Castro da~Silva.
\newblock From past to future: Rethinking eligibility traces.
\newblock \emph{arXiv}, 2312.12972, 2023.

\bibitem[Harb \& Precup(2016)Harb and Precup]{harb2016investigating}
Jean Harb and Doina Precup.
\newblock Investigating recurrence and eligibility traces in deep {Q}-networks.
\newblock In \emph{NeurIPS Deep Reinforcement Learning Workshop}, 2016.

\bibitem[Harutyunyan et~al.(2016)Harutyunyan, Bellemare, Stepleton, and Munos]{harutyunyan2016q}
Anna Harutyunyan, Marc~G. Bellemare, Tom Stepleton, and R{\'e}mi Munos.
\newblock {Q($\lambda$)} with off-policy corrections.
\newblock In \emph{International Conference on Algorithmic Learning Theory (ALT)}, 2016.

\bibitem[Kaelbling et~al.(1998)Kaelbling, Littman, and Cassandra]{kaelbling1998planning}
Leslie~Pack Kaelbling, Michael~L. Littman, and Anthony~R. Cassandra.
\newblock Planning and acting in partially observable stochastic domains.
\newblock \emph{Artificial Intelligence}, 101\penalty0 (1-2):\penalty0 99--134, 1998.

\bibitem[Kearns \& Singh(2000)Kearns and Singh]{kearns2000bias}
Michael~J. Kearns and Satinder Singh.
\newblock Bias-variance error bounds for temporal difference updates.
\newblock In \emph{Conference on Learning Theory (COLT)}, 2000.

\bibitem[Klopf(1972)]{klopf1972brain}
A.~Harry Klopf.
\newblock Brain function and adaptive systems: A heterostatic theory.
\newblock Technical report, Air Force Cambridge Research Laboratories, 1972.

\bibitem[Konidaris et~al.(2011)Konidaris, Niekum, and Thomas]{konidaris2011td_gamma}
George Konidaris, Scott Niekum, and Philip~S. Thomas.
\newblock $\text{TD}_\gamma$: Re-evaluating complex backups in temporal difference learning.
\newblock In \emph{Neural Information Processing Systems (NeurIPS)}, 2011.

\bibitem[Kozuno et~al.(2021)Kozuno, Tang, Rowland, Munos, Kapturowski, Dabney, Valko, and Abel]{kozuno2021revisiting}
Tadashi Kozuno, Yunhao Tang, Mark Rowland, R{\'e}mi Munos, Steven Kapturowski, Will Dabney, Michal Valko, and David Abel.
\newblock Revisiting {P}eng’s {Q}($\lambda$) for modern reinforcement learning.
\newblock In \emph{International Conference on Machine Learning (ICML)}, 2021.

\bibitem[Mahmood et~al.(2017)Mahmood, Yu, and Sutton]{mahmood2017multi}
A.~Rupam Mahmood, Huizhen Yu, and Richard~S. Sutton.
\newblock Multi-step off-policy learning without importance sampling ratios.
\newblock \emph{arXiv}, 1702.03006, 2017.

\bibitem[Minsky(1961)]{minsky1961steps}
Marvin~L. Minsky.
\newblock Steps toward artificial intelligence.
\newblock In \emph{Proceedings of the Institute of Radio Engineers (IRE)}, 1961.

\bibitem[Mousavi et~al.(2017)Mousavi, Schukat, Howley, and Mannion]{mousavi2017applying}
Seyed~Sajad Mousavi, Michael Schukat, Enda Howley, and Patrick Mannion.
\newblock Applying {Q($\lambda$)-Learning} in deep reinforcement learning to play {A}tari games.
\newblock In \emph{AAMAS Adaptive Learning Agents Workshop}, pp.\  1--6, 2017.

\bibitem[Munos et~al.(2016)Munos, Stepleton, Harutyunyan, and Bellemare]{munos2016safe}
R{\'e}mi Munos, Tom Stepleton, Anna Harutyunyan, and Marc~G. Bellemare.
\newblock Safe and efficient off-policy reinforcement learning.
\newblock In \emph{Neural Information Processing Systems (NeurIPS)}, 2016.

\bibitem[Robbins \& Monro(1951)Robbins and Monro]{robbins1951stochastic}
Herbert Robbins and Sutton Monro.
\newblock A stochastic approximation method.
\newblock \emph{The Annals of Mathematical Statistics}, 22\penalty0 (3):\penalty0 400--407, 1951.

\bibitem[Schulman et~al.(2015)Schulman, Moritz, Levine, Jordan, and Abbeel]{schulman2015high}
John Schulman, Philipp Moritz, Sergey Levine, Michael Jordan, and Pieter Abbeel.
\newblock High-dimensional continuous control using generalized advantage estimation.
\newblock In \emph{International Conference on Learning Representations (ICLR)}, 2015.

\bibitem[Sutton(1984)]{sutton1984temporal}
Richard~S. Sutton.
\newblock \emph{Temporal Credit Assignment in Reinforcement Learning}.
\newblock PhD thesis, University of Massachusetts Amherst, 1984.

\bibitem[Sutton(1988)]{sutton1988learning}
Richard~S. Sutton.
\newblock Learning to predict by the methods of temporal differences.
\newblock \emph{Machine Learning}, 3\penalty0 (1):\penalty0 9--44, 1988.

\bibitem[Sutton \& Barto(2018)Sutton and Barto]{sutton2018reinforcement}
Richard~S. Sutton and Andrew~G. Barto.
\newblock \emph{Reinforcement Learning: An Introduction}.
\newblock MIT Press, 2nd edition, 2018.

\bibitem[Tang et~al.(2024)Tang, Rowland, Munos, Pires, and Dabney]{tang2024off}
Yunhao Tang, Mark Rowland, R{\'e}mi Munos, Bernardo~{\'A}vila Pires, and Will Dabney.
\newblock Off-policy distributional {Q($\lambda$)}: Distributional {RL} without importance sampling.
\newblock \emph{arXiv}, 2402.05766, 2024.

\bibitem[Thomas et~al.(2015)Thomas, Niekum, Theocharous, and Konidaris]{thomas2015policy}
Philip~S. Thomas, Scott Niekum, Georgios Theocharous, and George Konidaris.
\newblock Policy evaluation using the {$\Omega$}-return.
\newblock \emph{Neural Information Processing Systems (NeurIPS)}, 2015.

\bibitem[Tsitsiklis \& Van~Roy(1997)Tsitsiklis and Van~Roy]{tsitsiklis1997analysis}
John~N. Tsitsiklis and Benjamin Van~Roy.
\newblock An analysis of temporal-difference learning with function approximation.
\newblock \emph{IEEE Transactions on Automatic Control}, 42\penalty0 (5):\penalty0 674--690, 1997.

\bibitem[van Seijen(2016)]{vanseijen2016effective}
Harm van Seijen.
\newblock Effective multi-step temporal-difference learning for non-linear function approximation.
\newblock \emph{arXiv}, 1608.05151, 2016.

\bibitem[Watkins(1989)]{watkins1989learning}
Christopher J. C.~H. Watkins.
\newblock \emph{Learning from Delayed Rewards}.
\newblock PhD thesis, University of Cambridge, 1989.

\bibitem[White \& White(2016)White and White]{white2016greedy}
Martha White and Adam White.
\newblock A greedy approach to adapting the trace parameter for temporal difference learning.
\newblock In \emph{International Conference on Autonomous Agents and Multiagent Systems (AAMAS)}, pp.\  557--565, 2016.

\bibitem[Yu(2012)]{yu2012least}
Huizhen Yu.
\newblock Least squares temporal difference methods: An analysis under general conditions.
\newblock \emph{SIAM Journal on Control and Optimization}, 50\penalty0 (6):\penalty0 3310--3343, 2012.

\end{thebibliography}
